\pdfoutput=1

\documentclass[11pt]{article}

\usepackage[]{acl}

\usepackage{times}
\usepackage{latexsym}

\usepackage[T1]{fontenc}

\usepackage[utf8]{inputenc}

\usepackage{CJK}
\usepackage{tabularx}
\usepackage{graphicx}
\usepackage{booktabs}
\usepackage{amsmath}
\usepackage{algorithm}
\usepackage{algorithmic}
\usepackage{float}
\usepackage{color}
\usepackage{arydshln}
\usepackage{bm}
\usepackage{amssymb}
\usepackage{multirow}
\usepackage{multicol}
\usepackage{bbding}
\usepackage{arydshln}
\usepackage{makecell}

\usepackage{microtype}

\title{Zero-shot Cross-lingual Conversational Semantic Role Labeling}

\author{
Han Wu$^{1,2}$, Haochen Tan$^{1,2}$, Kun Xu$^{3}\thanks{~~The paper was done when working in Tencent AI Lab.}$, Shuqi Liu$^{1,2}$, Lianwei Wu$^{4}$, Linqi Song$^{1,2}$\\
     $^{1}$City University of Hong Kong Shenzhen Research Institute \\
     $^{2}$Department of Computer Science, City University of Hong Kong, $^{3}$Huawei\\
	 $^{4}$National Engineering Laboratory for Integrated Aero-Space-Ground-Ocean Big Data Application Technology, \\School of Computer Science, Northwestern Polytechnical University\\
	 \texttt{\{hanwu32-c,haochetan-2,shuqiliu-4\}@my.cityu.edu.hk}\\
	 \texttt{syxu828@gmail.com}, \texttt{wlw@nwpu.edu.cn} \\
	 \texttt{linqi.song@cityu.edu.hk}
}

\begin{document}
\maketitle
\begin{CJK*}{UTF8}{gbsn}
\begin{abstract}
While conversational semantic role labeling (CSRL) has shown its usefulness on Chinese conversational tasks, it is still under-explored in non-Chinese languages due to the lack of multilingual CSRL annotations for the parser training.
To avoid expensive data collection and error-propagation of translation-based methods, we present a simple but effective approach to perform zero-shot cross-lingual CSRL.
Our model implicitly learns language-agnostic, conversational structure-aware and semantically rich representations with the hierarchical encoders and elaborately designed pre-training objectives.
Experimental results show that our model outperforms all baselines by large margins on two newly collected English CSRL test sets.
More importantly, we confirm the usefulness of CSRL to non-Chinese conversational tasks such as the question-in-context rewriting task in English and the multi-turn dialogue response generation tasks in English, German and Japanese by incorporating the CSRL information into the downstream conversation-based models. We believe this finding is significant and will facilitate the research of non-Chinese dialogue tasks which suffer the problems of ellipsis and anaphora.
\end{abstract}

\section{Introduction}
\begin{figure}[t!]
    \centering
    \includegraphics[width=1.0\linewidth]{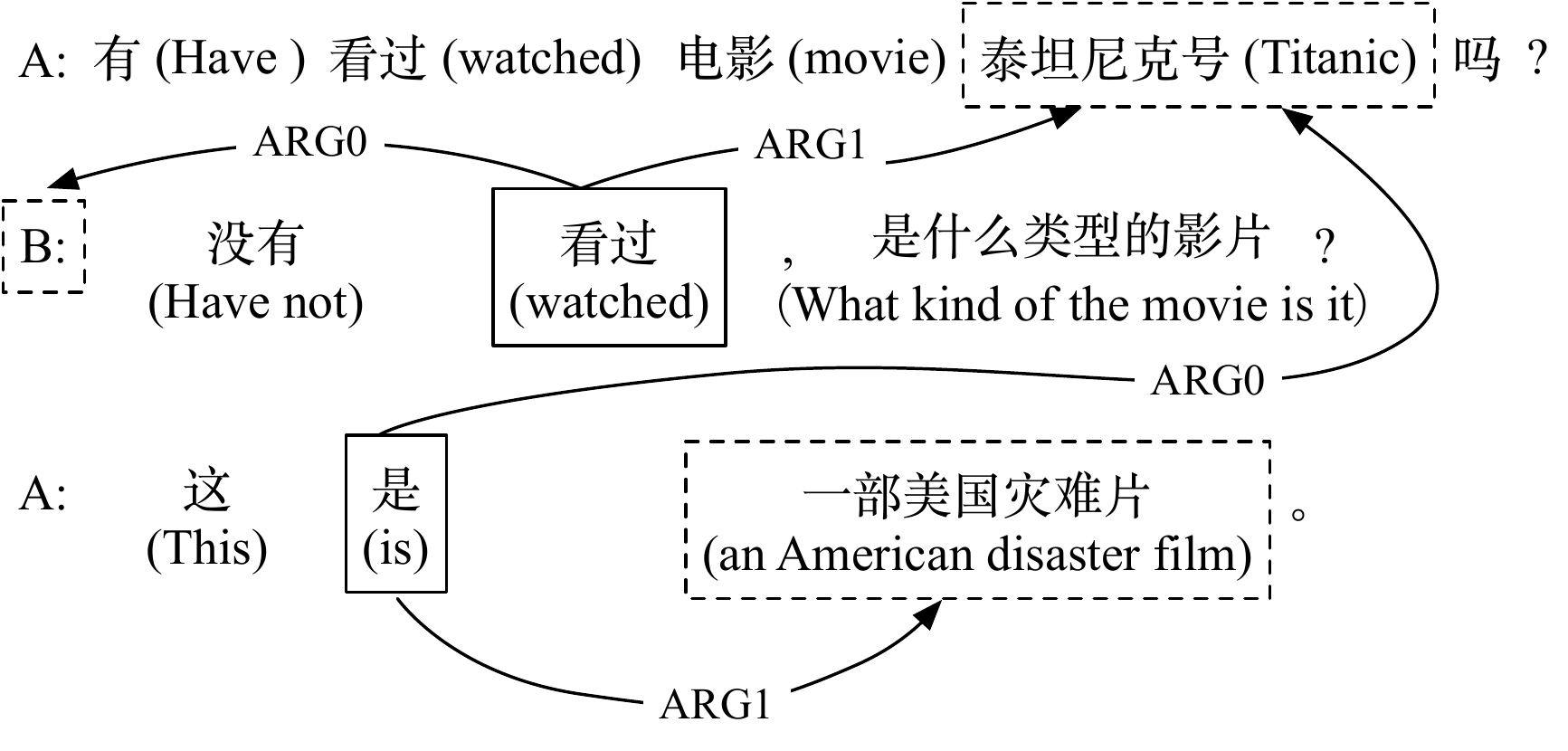}
    \caption{An example of CSRL parsing.}
    \label{fig:example}
    \vspace{-1.5em}
\end{figure}

Conversational Semantic Role Labeling (CSRL) \citep{xu2021conversational} is a recently proposed dialogue understanding task, which aims to extract predicate-argument pairs from the entire conversation.
Figure \ref{fig:example} illustrates a CSRL example where a CSRL parser is required to identify ``{\small《泰坦尼克号》}(Titanic)'' as the \texttt{ARG1} argument of the predicate ``{\small 看过} (watched)" and the \texttt{ARG0} argument of the predicate ``{\small 是} (is)".
We can see that in the original conversation, ``{\small《泰坦尼克号》}(Titanic)'' is omitted in the second turn and referred as ``{\small 这} (this)" in the last turn.
By recovering the dropped and referred components in conversation, CSRL has shown its usefulness to a set of Chinese dialogue tasks, including multi-turn dialogue rewriting \citep{su-etal-2019-improving} and response generation \citep{wu-etal-2019-proactive}.
However, there remains a paucity of evidence on its effectiveness towards non-Chinese languages owing to the lack of multilingual CSRL models.
To adapt a model into new languages, previous solutions can be divided into three categories: 1) manually annotating a new dataset in the target language \citep{daza2020x} 2) borrowing machine translation and word alignment techniques to transfer the dataset from the source language into the target language \citep{daza2019translate, fei2020cross-acl} 3) zero-shot transfer learning with multilingual pre-trained language model \citep{rijhwani2019zero, sherborne2021zero}. Due to the fact that manually collecting annotations is costly and translation-based methods might introduce translation or word alignment errors, zero-shot cross-lingual transfer learning is more practical to the NLP community.

Recent works have witnessed prominent performances of multilingual pre-trained language models (PrLMs) \citep{devlin2019bert, conneau2020unsupervised} on cross-lingual tasks, including machine translation \citep{lin2020pre, chen-etal-2021-zero}, semantic role labeling (SRL) \citep{conia2020bridging, conia2021unifying} and semantic parsing \citep{fei2020cross-taslp, sherborne2021zero}.
However, cross-lingual CSRL, as a combination of three challenging tasks (i.e., cross-lingual task, dialogue task and SRL task), suffers three outstanding difficulties: 1) \textbf{latent space alignment} - how to map word representations of different languages into an overlapping space; 2) \textbf{conversation structure encoding} - how to capture high-level dialogue features such as speaker dependency and temporal dependency; and 3) \textbf{semantic arguments identification} - how to highlight the relations between the predicate and its arguments, wherein PrLMs can only partially encode multilingual inputs to an overlapping vector space.
Although there are some success that can separately achieve structural conversation encoding \citep{mehri-etal-2019-pretraining, zhang-zhao-2021-structural} and semantic arguments identification \citep{wu-etal-2021-domain}, a unified method for jointly solving these problems is still under-explored, especially in a cross-lingual scenario.

In this work, we summarize our contributions as follows:
    \textbf{(1)} We propose a simple but effective model which consists of three modules, namely cross-lingual language model (CLM), structure-aware conversation encoder (SA-Encoder) and predicate-argument encoder (PA-Encoder), and five well-designed pre-training objectives. Our model implicitly learns language-agnostic, conversational structure-aware and semantically rich representations to perform zero-shot cross-lingual CSRL.
    \textbf{(2)} Experiments show that our method achieves impressive cross-lingual performance on the language pair (Zh$\rightarrow$En) , and outperforms all baselines on the two newly collected English CSRL test sets.
    \textbf{(3)} We confirm the usefulness of CSRL to the question-in-context rewriting task in English and multi-turn response generation tasks in English, German and Japanese.
    We believe this finding is important and will facilitate the research of non-Chinese dialogue tasks that suffer from ellipsis and anaphora.
    \textbf{(4)} We release our code, the new annotated English CSRL test sets and checkpoints of our best models to facilitate the further research at \url{https://github.com/hahahawu/Zero-Shot-XCSRL}.

\section{Related Work}
\paragraph{Zero-shot cross-lingual transfer learning.} Recently, thanks to the rapid development of multilingual pre-trained language models such as multilingual BERT \citep{devlin2019bert} and XLM-R \citep{conneau2020unsupervised}, a number of approaches have been proposed for zero-shot cross-lingual transfer learning on various downstream tasks, including semantic parsing \citep{sherborne2021zero}, natural language generation \citep{shen2018zero} and understanding \citep{liu2019zero, lauscher2020zero}.
In this work, we claim our method is zero-shot because no non-Chinese CSRL annotations are seen during the CSRL training stage. For decoding, we directly use the cross-lingual CSRL model trained on Chinese CSRL data to analyze conversations in other languages.
To our best knowledge, our work is the first step to cross-lingual CSRL.

\paragraph{Conversational semantic role labeling.} While ellipsis and anaphora frequently occur in dialogues, \citet{xu2021conversational} observed that most of the dropped or referred components can be found in dialogue histories. Following this observation, they proposed conversational semantic role labeling (CSRL) which required the model to find predicate-argument structures over the entire conversation instead of a single sentence. In this way, when analyzing a predicate in the latest utterance, a CSRL model needs to consider both the current turn and previous turns to search potential arguments, and thus might recover the omitted components.
Furthermore, \citet{xu2020semantic, xu2021conversational} also confirmed the usefulness of CSRL to Chinese dialogue tasks by applying CSRL information into downstream dialogue tasks. However, there are still two main problems to be solved for CSRL task: (1) the performance of current state-of-the-art CSRL model \citep{xu2021conversational} is still far from satisfactory due to the lack of high-level conversational and semantic features modeling; (2) the usefulness of CSRL to conversational tasks in non-Chinese languages has not been confirmed yet due to the lack of cross-lingual CSRL models.
In this work, we primarily focus on the latter problem and propose a simple but effective model to perform cross-lingual CSRL.
We would like to distinguish our work from the work \citep{wu-etal-2021-csagn} which purely focuses on improving the monolingual CSRL performance where they try to model predicate-aware representations. This solution could benefit to monolingual CSRL task, but hurt the cross-lingual performance, because the relative positions of the predicates may differ from language to language.

\begin{figure*}[t!]
    \centering
    \includegraphics[width=0.98\linewidth]{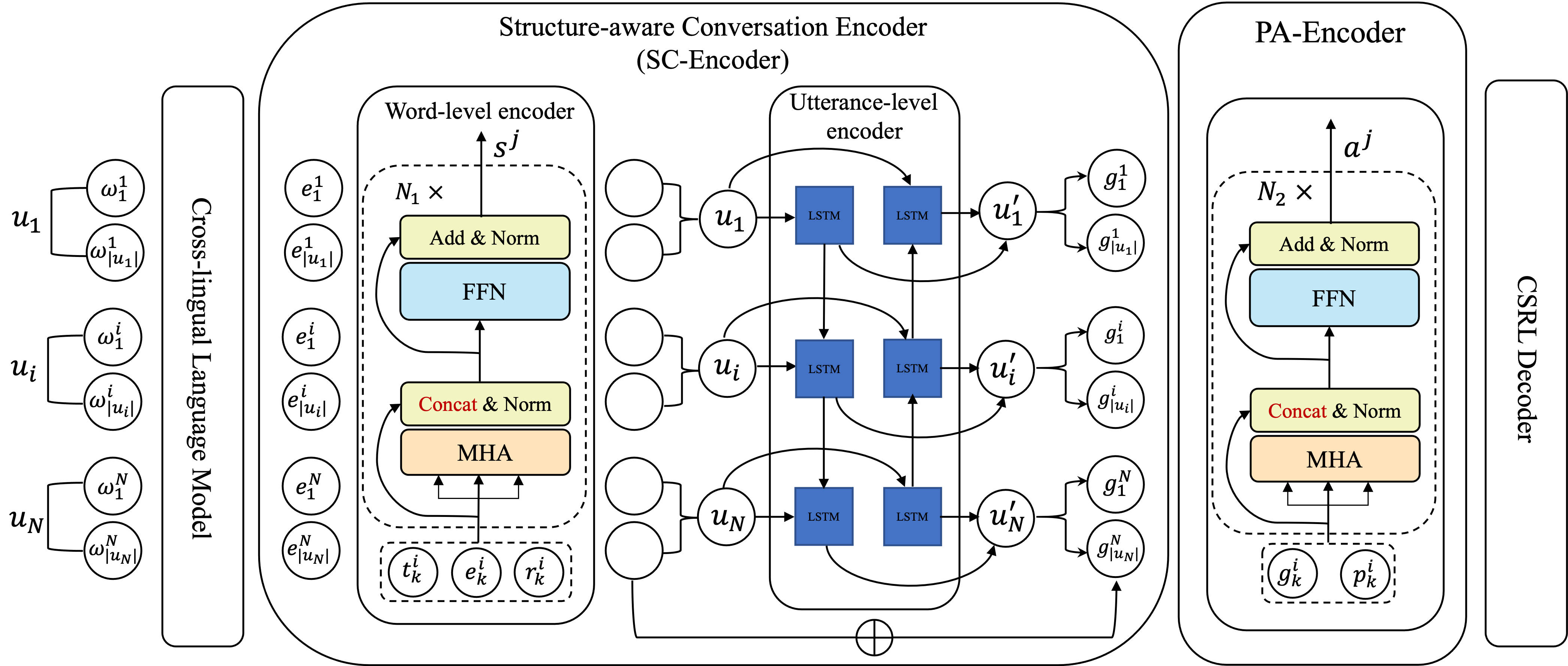}
    \caption{Overall model architecture.}
    \label{fig:model}
    \vspace{-0.2cm}
\end{figure*}

\section{Methodology}
Following \citet{xu2021conversational}, we solve the CSRL task as a sequence labeling problem. Formally, given a dialogue $C = \{u_1, u_2, ..., u_N\}$ of $N$ utterances, where $u_i = \{w^i_1, w^i_2, ..., w^i_{|u_i|}\}$ consisting of a sequence of words, and a predicate indicator ${p} = ({p}_1^1, ..., {p}_k^i, ..., {p}_{|u_N|}^N)$ used to identify whether a word is the predicate or not, our goal is to assign each word with a semantic role label $l \in L$ where $L$ is the label set. We also incorporate speaker role indicator ${r}$ to distinguish speakers, and dialogue turn indicator ${t}$ to distinguish dialogue turns.

\subsection{Architecture} \label{sec:architecture}
\paragraph{Cross-lingual Language Model (CLM)}
We concatenate all utterances into a sequence and then use a pre-trained cross-lingual language model such as XLM-R \citep{conneau2020unsupervised} or mBERT \citep{devlin2019bert} to capture the syntactic and semantic characteristics. Following \citet{conia2021unifying}, we obtain word representations $\textbf{e} \in \mathbb{R}^{|S| \times d}$ by concatenating the hidden states of the four top-most layers of the language model, where $|S|$ is the sequence length and $d$ is the dimension of the hidden state.

\paragraph{Structure-aware Conversation Encoder (SC-Encoder)}
Different from standard SRL\citep{carreras2005introduction}, CSRL requires the models to find arguments from not only the current turn, but also previous turns, thus bringing more challenges of dialogue modeling. To address this problem, we propose a universal structure-aware conversation encoder which comprises of two parts, i.e., word-level encoder and utterance-level encoder. 
Formally, with the speaker role embedding $\bm{r} \in \mathbb{R}^{|S| \times d}$ and dialogue turn embedding $\bm{t} \in \mathbb{R}^{|S| \times d}$, the word-level encoder computes a sequence of timestep encodings $\bm{s} \in \mathbb{R}^{|S| \times d}$ as follows:
\begin{equation}
\small
    \bm{s}_{(i, k)}^j = \left\{
    \begin{array}{ll}
        \bm{e}_k^i \oplus \bm{t}_k^i \oplus \bm{r}_k^i & \text{if}~~j=0 \\
        \bm{s}_{(i, k)}^{j-1} \oplus \textsc{MTrans}^j(\bm{s}_{(i, k)}^{j-1}) & \text{otherwise}
    \end{array}\right.
\end{equation}
where $\bm{s}_{(i, k)}^j$ is the timestep encoding of $k$-th token in $i$-th utterance from $j$-th word-level encoder layer while $j \in (0, \dots, N_1)$, $\oplus$ represents vector concatenation, and \textsc{MTrans} is the \textbf{M}odified \textbf{Trans}former encoder layer. Concretely, we replace the \texttt{[Add]} operation in the first residual connection layer with \texttt{[Concat]} because we argue that concatenation is a superior approach to preserve the information from previous layers\footnote{More details about \textsc{MTrans} in Appendix \ref{apx:mtrans}.}.

We obtain utterance representations $\bm{u} \in \mathbb{R}^{N \times d}$ by max-pooling over words in the same utterance. Then we pass the resulting utterance representations $\bm{u}$ through a stack of Bi-LSTM \citep{hochreiter1997long} layers to obtain the sequentially encoded utterance representations $\bm{u}^{\prime} \in \mathbb{R}^{N \times d}$. Finally, we combine the utterance-level feature $\bm{u}^{\prime}$ with the word-level feature $\bm{s}$ to obtain structure-aware dialogue context representations $\bm{g} \in \mathbb{R}^{|S| \times d}$  as follows:
\begin{equation}
    \bm{g}_k^i = \text{Swish}(\mathbf{W}^g[\bm{s}_{(i, k)}^{N_1} \oplus \bm{u}_i^{\prime}] + \mathbf{b}^g)
\end{equation}
where $\text{Swish}(x)=x\cdot\text{sigmoid}(x)$ is a non-linear activation function, $\bm{s}_{i,k}^{N_1}$ is the encoding of $k$-th token in $i$-th utterance from the last layer of the word-level encoder. $\mathbf{W}^g$ and $\mathbf{b}^g$ are trainable parameters.

\begin{figure}[t!]
    \centering
    \includegraphics[width=1.0\linewidth]{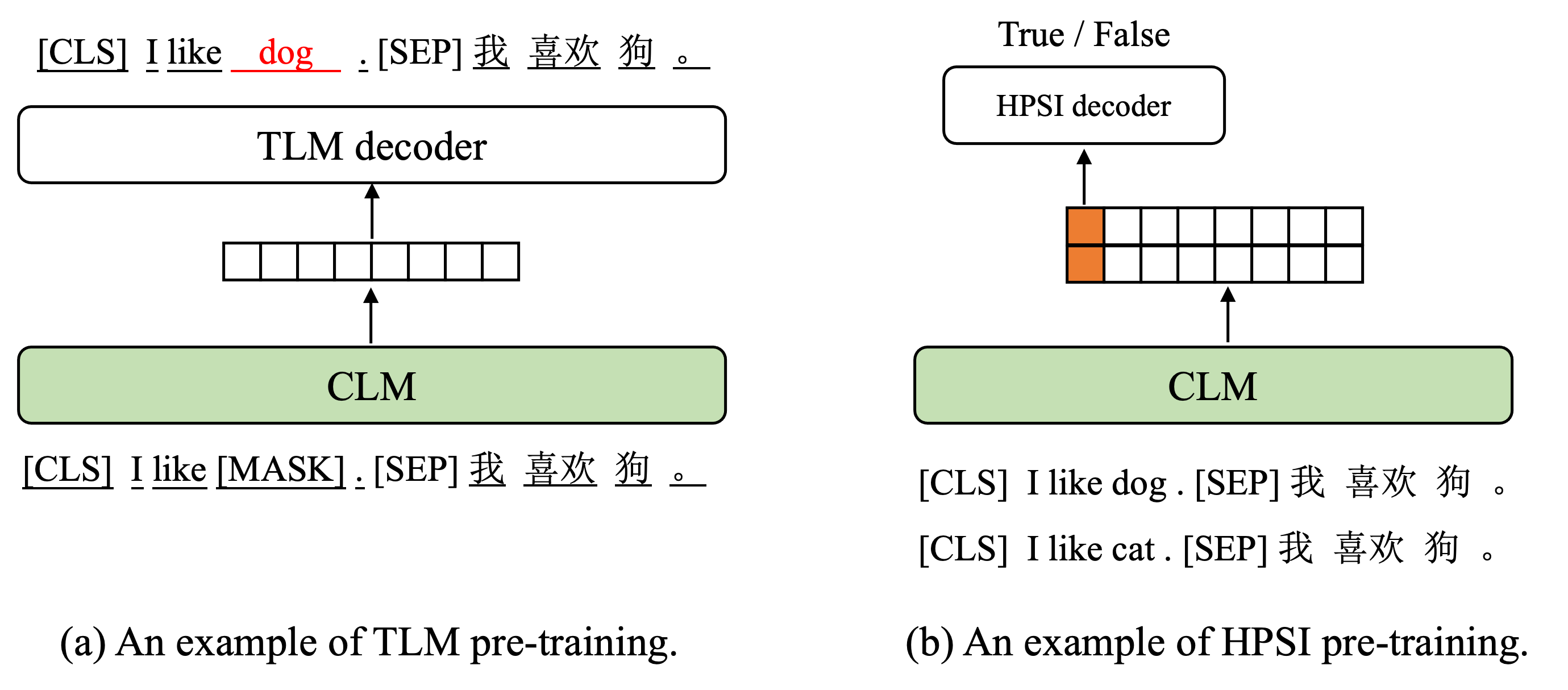}
    \caption{Illustration of pre-training objectives for latent space alignment.}
    \label{fig:latent_space_alignment}
    \vspace{-0.2cm}
\end{figure}

\begin{figure}[t!]
    \centering
    \includegraphics[width=1.0\linewidth]{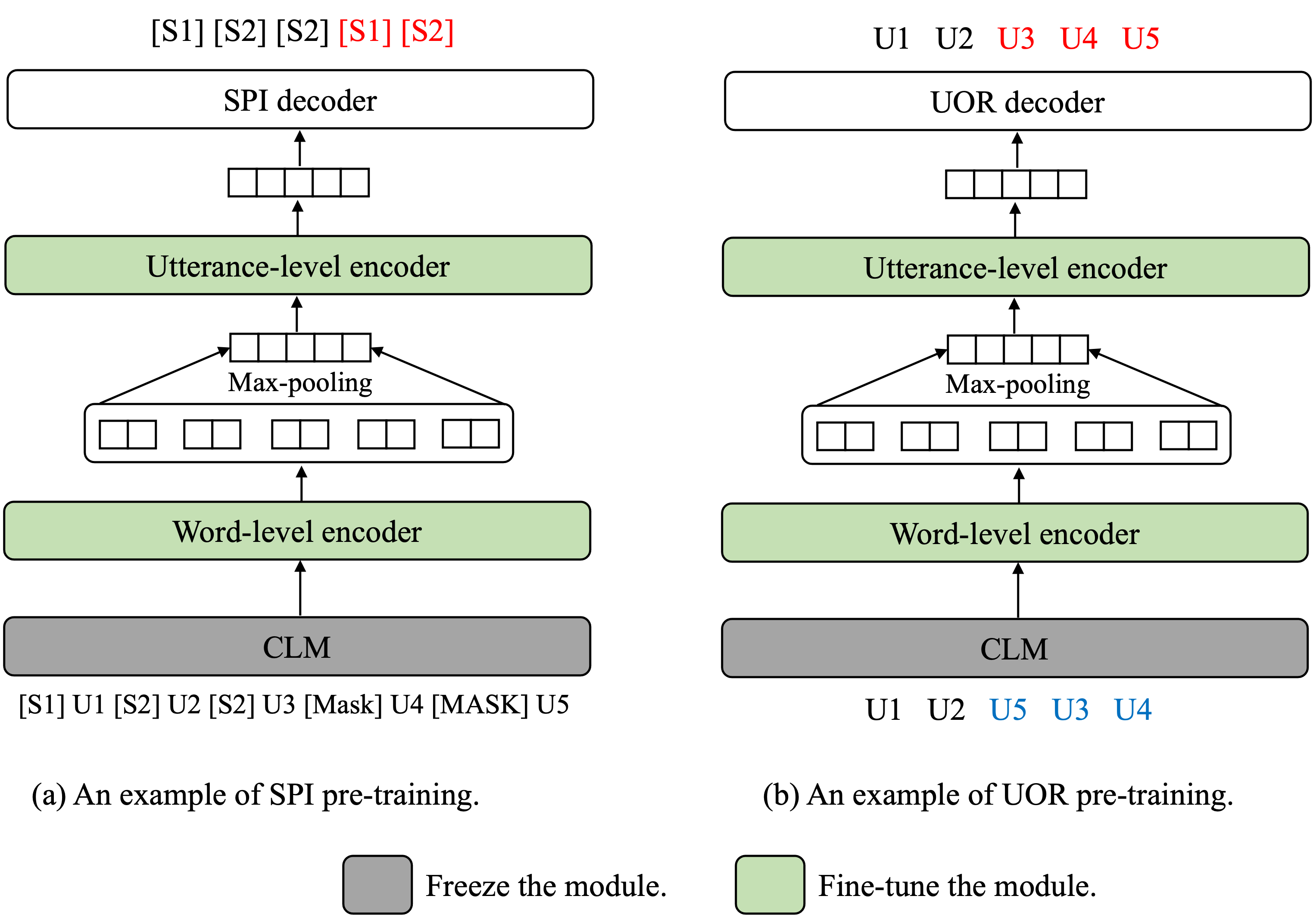}
    \caption{Illustration of pre-training objectives for conversation structure encoding.}
    \label{fig:conversation_structure_encoding}
    \vspace{-0.5cm}
\end{figure}

\paragraph{Predicate-Argument Encoder (PA-Encoder)}
We introduce the third module (i.e., predicate-argument encoder) whose goal is to capture the relations between each predicate-argument couple that appears in the conversation. Similar with the word-level encoder, we use a stack of $\textsc{MTrans}$ layers to implement this encoder. Formally, with the predicate embedding $\bm{p} \in \mathbb{R}^{|S| \times d}$, the model calculates the predicate-specific argument encodings $\bm{a} \in \mathbb{R}^{|S| \times d}$ as follows:
\begin{equation}
\small
    \bm{a}_{(i, k)}^j = \left\{
    \begin{array}{ll}
        \bm{g}_k^i \oplus \bm{p}_k^i & \text{if}~~j=0 \\
        \bm{a}_{(i, k)}^{j-1} \oplus \textsc{MTrans}^j(\bm{a}_{(i, k)}^{j-1}) & \text{otherwise}
    \end{array}\right.
\end{equation}
where $\bm{a}_{(i,k)}^{j}$ is the argument encoding of $k$-th token in $i$-th utterance from $j$-th encoder layer while $j \in (0,\dots, N_2)$. Finally, we obtain the semantic role encoding $\bm{l}$ using the resulting argument encodings from the last layer of the predicate-argument encoder:
\begin{equation}
    \bm{l}_k^i = \text{Swish}(\mathbf{W}^l\bm{a}_{(i,k)}^{N_2} + \mathbf{b}^l)
\end{equation}
In particular, our proposed model is mostly language-agnostic since we do not explicitly introduce any language-specific knowledge such as word order, part-of-speech tags or dependent relations, and only introduce the predicate indicator that might contain some language-specific information in the semantic module, which would not affect latent space alignment and dialogue modeling.

\begin{figure}[t!]
    \centering
    \includegraphics[width=0.8\linewidth]{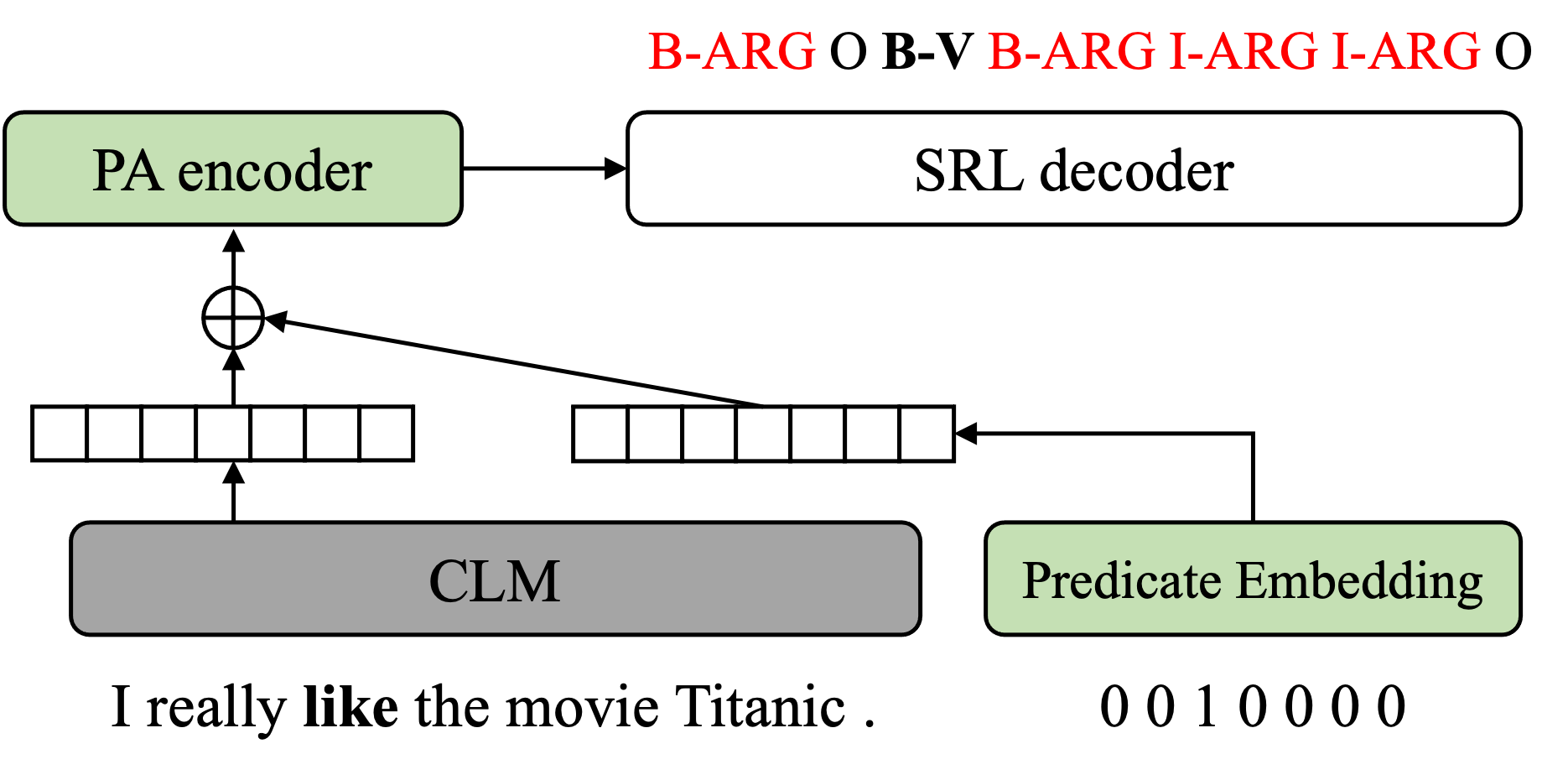}
    \caption{Illustration of pre-training objectives for semantic argument identification.}
    \label{fig:semantic_argument_identification}
    \vspace{-1em}
\end{figure}

\subsection{Pre-training Objectives} \label{sec:pretrain-obj}
Besides the universal model, we also elaborately design five pre-training objectives to model task-specific but language-agnostic features for better cross-lingual performance.
In this section, we divide our pre-training objectives into three groups according to the challenges to be solved.

\paragraph{Latent space alignment}
In cross-lingual language module, we use mBERT or XLM-R to align the latent space of different languages. Although mBERT and XLM-R have exhibited good alignment ability, even both of them are trained with unpaired data, we may further improve it when we have access to parallel data.

We first use translation language model (TLM) \citep{conneau2019cross} to learn word-level alignment ability. Concretely, we concatenate parallel sentences as a single consecutive token sequence with special tokens separating them and then perform masked language modeling (MLM) \citep{devlin2019bert} on the concatenated sequence.
Besides, we also attempt to improve sentence-level alignment ability using hard parallel sentence identification (HPSI). Specifically, we select a pair of parallel or non-parallel sentences from the training set with equal probability. Then the model is required to predict whether the sampled sentence pair is parallel or not.
Different from the standard PSI \citep{dou2021word}, we sample the non-parallel sentence upon the n-gram similarity or construct it by text perturbation (details in Appendix \ref{apx:hpsi}) instead of in a random manner.
Figure \ref{fig:latent_space_alignment} illustrates the workflows of TLM and HPSI. We pre-train the CLM using the combination of TLM and HPSI, finally achieving latent space alignment.

\begin{table*}[t!]
\small
    \fontsize{10}{11} \selectfont
    \setlength\tabcolsep{2.1pt}
    \centering
    \bgroup
    \def\arraystretch{1.2}
    \begin{tabular}{l|cccccc}
    \toprule
        Dataset & language & \#dialogue & \#utterance & \#predicate & \#tokens per utterance & cross ratio \\
        \hline
        DuConv & ZH & 3,000 & 27,198 & 33,673 & 10.56 & 21.89\% \\
        Persona-Chat & EN & 50 & 2,669 & 477 & 17.96 & 17.74\% \\
        CMU-DoG & EN & 50 & 3,217 & 450 & 12.57 & 7.41\% \\
    \bottomrule
    \end{tabular}
    \egroup
    \caption{Statistics of the annotations on DuConv, NewsDialog and PersonalDialog.}
    \label{tab:statistics}
    \vspace{-1em}
\end{table*}

\paragraph{Conversation structure encoding}
Although there are a number of pre-training objectives proposed to learn dialogue context representations \citep{mehri-etal-2019-pretraining} and structural representations \citep{zhang-zhao-2021-structural}, we tend to explicitly model speaker dependency and temporal dependency in the conversation, both of which have been proven to be critical to CSRL task \citep{xu2021conversational}.

We first propose speaker role identification (SPI) to learn speaker dependency in the conversation. Specifically, we randomly sample $K_1\%$ utterances and replace their speaker indicators with special mask tags. To make the task harder and effective, we split the utterances into clauses if only two interlocutors utter in turn in a conversation. The goal of SPI is to predict the masked speaker roles according to the corrupted speaker indicators and context.
Secondly, we borrow utterance order permutation (UOR) to encourage the model to be aware of temporal connections among utterances in the context.
Concretely, given a set of utterances, we randomly shuffle the last $K_2\%$ utterances and require the model to organize them into a coherent context.
Figure \ref{fig:conversation_structure_encoding} illustrates the workflows of SPI and UOR. We pre-train the SC-Encoder using the combination of SPI and UOR.

\paragraph{Semantic arguments identification}
The core of all SRL-related tasks is to recognize the predicate-argument pairs from the input.
Therefore, we propose semantic arguments identification (SAI) objective to strengthen the correlations between the predicate and its arguments with the help of external standard SRL corpus, i.e., CoNLL-2012.
Specifically, for each SRL sample, we only focus on those arguments, including ARG0-4, ARG-LOC, ARG-TMP and ARG-PRP, all of which are defined in both SRL and CSRL tasks. The model is encouraged to find the textual spans of these arguments with the given predicate. We believe this objective would benefit to boundary detection, especially for location and temporal arguments.
Figure \ref{fig:semantic_argument_identification} illustrates the workflow of SAI. We drop the SC-Encoder to fit in standard SRL samples which do not have any conversational characteristics.

\subsection{Training}

\paragraph{Hierarchical Pre-training}
The pre-training is hierarchically conducted according to different modules, and the pre-training of the upper module is based on the pre-trained lower modules.
Specifically, we first train CLM module with TLM and HPSI; then we train SC-Encoder with SPI and UOR while keeping the weights of pre-trained CLM module unchanged; finally we train PA-Encoder with SAI while freezing the weights of pre-trained CLM and SC-Encoder modules.
Hopefully, we expect that each module could acquire different knowledge with specific pre-training objectives.

\paragraph{CSRL training}
We initialize the specific modules, including CLM, SC-Encoder and PA-Encoder, from the pre-trained checkpoints.
The CSRL model is trained only using Chinese CSRL annotations and no additional data is introduced during the CSRL training stage.
We train our model to minimize the cross-entropy error for a training sample with label $y$ based on the semantic role encoding $\bm{l}$,
\vspace{-1em}
\begin{equation}
    p = \text{softmax}(\bm{l}_t)\quad \mathcal{L}_{CSRL} = -\sum_{l=1}^L y\log p
\end{equation}
\section{Experiments}

\begin{table*}[t!]
\fontsize{9}{10} \selectfont
\setlength\tabcolsep{4.5pt}
\centering
\bgroup
\def\arraystretch{1.2}
\begin{tabular}{lclcccccccccccc}
\toprule[0.8pt]
\multirow{2}{*}{Method} & Trainable & \multicolumn{3}{c}{DuConv} & & \multicolumn{3}{c}{Persona-Chat} & & \multicolumn{3}{c}{CMU-DoG} \\
\cline{3-5} \cline{7-9} \cline{11-13}
& parameters & F1$_{all}$ & F1$_{cross}$ & F1$_{intra}$ & & F1$_{all}$ & F1$_{cross}$ & F1$_{intra}$ & & F1$_{all}$ & F1$_{cross}$ & F1$_{intra}$ \\
\hline
SimpleBERT & 117 M & 86.54 & 81.62 & 87.02 & & - & - & - & & - & - & - \\
CSRL-BERT & 147 M & 88.46 & 81.94 & 89.46 & & - & - & - & & - & - & - \\
CSAGN & 163 M & {\textbf{89.47}} & {\textbf{84.57}} & {\textbf{90.15}} & & - & - & - & & - & - & - \\
SimpleXLMR & 292 M & 84.75 & 63.44 & 85.12 & & 62.96 & 14.29 & 63.03 & & 50.54 & 14.29 & 58.50 \\
CSRL-XLMR & 320 M & 88.03 & 78.12 & 89.33 & & 63.18 & 18.71 & 65.05 & & 53.84 & 34.20 & 59.78 \\
CSAGN-XLMR & 338 M & 88.52 & 82.45 & 89.98 & & 63.02 & 17.82 & 64.97 & & 52.73 & 30.11 & 58.91 \\
Translation-test & - & - & - & - & & 63.49 & 13.90 & 66.67 & & 47.91 & 27.44 & 50.92\\
Translation-train & - & - & - & - & & 60.12 & 9.67 & 62.50 & & 44.27 & 25.40 & 47.87\\
\hline
\multicolumn{4}{l}{\textit{Fine-tune all parameters}} \\
\hline
Ours$_\text{mBERT}$ & 272 M & 87.20 & 81.14 & 88.11 & & 58.38 & 9.39 & 61.77 & & 48.13 & 20.92 & 52.91 \\
Ours$_{\text{XLM-R}}$ & 372 M & 88.35 & 83.39 & 89.21 & & \textbf{67.29} & 24.29 & \textbf{70.61} & & \textbf{61.74} & \textbf{60.32} & \textbf{62.67}\\
Ours$_{\text{XLM-R + pre-train}}$ & 372 M & {88.60} & {84.10} & {89.24} & & 67.23 & \textbf{25.43} & 69.89 & & 59.24 & 58.94 & 60.89 \\
\hline
\multicolumn{4}{l}{\textit{Freeze parameters of the language model}} \\
\hline
Ours$_{\text{mBERT}}$ & 180 M & 87.08 & 81.46 & 87.98 & & 59.04 & 11.23 & 62.13 & & 48.87 & 21.78 & 53.54\\
Ours$_{\text{XLM-R}}$ & 180 M & 88.30 & 83.38 & 89.17 & & 65.57 & 24.11 & 68.51 & & \textbf{59.60} & 56.16 & \textbf{60.78}\\
Ours$_{\text{XLM-R + pre-train}}$ & 180 M & {88.60} & {83.72} & {89.27} & & \textbf{66.75} & \textbf{24.13} & \textbf{69.44} & & 58.45 & \textbf{58.92} & 58.82\\
\toprule[0.8pt]
\end{tabular}
\egroup
\caption{Evaluation results on the DuConv, Persona-Chat and CMU-DoG datasets.}
\label{tab:csrl-results}
\vspace{-0.5cm}
\end{table*}
\subsection{Datasets}
\paragraph{CSRL data}
We use the same split as \citet{xu2021conversational} where DuConv annotations are splitted into 80\%/10\%/10\% as train/dev/in-domain test set.
Furthermore, we manually collect two CSRL test sets for cross-lingual evaluation based on Persona-Chat\citep{zhang2018personalizing} and CMU-DoG\citep{zhou2018dataset}, both of which are English conversation datasets. The CSRL data annotation is difficult because it needs great expertise in SRL and dialogue. So we only explore cross-lingual CSRL on Chinese$\rightarrow$English (Zh$\rightarrow$En) here, and we leave other languages for future work.

Following the instructions in \citet{xu2021conversational}, we manually collect two out-of-domain CSRL test sets based on English dialogue datasets Persona-Chat and CMU-DoG. Specifically, we also annotate the arguments ARG0-4, ARG-TMP, ARG-LOC and ARG-PRP and require that the labeled arguments can only appear in the current turn or the previous turns.
We employ three annotators who have studied Chinese CSRL annotations for a period time before this annotation. The first two annotators are required to label all cases and any disagreements between them are solved by the third annotator.
The statistics of the datasets are listed in Table \ref{tab:statistics}.

\paragraph{Pre-training data}
For TLM and HPSI objectives which requires parallel data to enhance alignment ability, we choose IWSLT'14 English$\leftrightarrow$Chinese (En$\leftrightarrow$Zh) translations\footnote{https://wit3.fbk.eu/}. For SPI and UOR objectives whose goal is to model high-level conversational features, we select samples from Chinese conversation dataset (i.e., DuConv) and English conversation datasets (i.e., Persona-Chat and CMU-DoG) with equal probability. For SAI, we borrow the Chinese and English SRL annotations from CoNLL-2012\citep{pradhan2012conll}.

We stress that by \textbf{keeping the sampling balance} of Chinese and English data for every pre-training objective and \textbf{sharing all parameters across the languages}, our model would capture task-specific but language-agnostic features.

\subsection{Experimental Setup}
Following previous work\citep{xu2021conversational}, we evaluate our system on micro-average F1$_{all}$, F1$_{cross}$ and F1$_{intra}$ over the (predicate, argument, label) tuples, wherein we calculate F1$_{cross}$ and F1$_{intra}$ over the arguments in the different, or same turn as the predicate. We refer these two types of arguments as \emph{cross}-arguments and \emph{intra}-arguments.
For language in-domain evaluation, we compare to \emph{SimpleBERT} \citep{shi2019simple}, \emph{CSRL-BERT} \citep{xu2021conversational} and \emph{CSAGN} \citep{wu-etal-2021-csagn}, all of which employ the Chinese pre-trained language model as the backbone. For cross-lingual evaluation, we compare to \emph{SimpleXLMR}, \emph{CSRL-XLMR} and \emph{CSAGN-XLMR} by simply replacing the BERT backbones of those models with XLM-R. Additionally, we also compare to the back-translation baselines, i.e., Translate-test and Translate-train. Specifically, Translate-test means that the English test data is translated and projected to Chinese annotations using Google Translate \citep{wu2016google} and the state-of-the-art word alignment toolkit Awesome-align\citep{dou2021word}. Similarly, Translate-train means the Chinese training data is translated and projected to English annotations for training. We feed the translated samples into CSAGN/CSAGN-XLMR to obtain the back-translation results.

\begin{table*}[t!]
\fontsize{9}{10} \selectfont
\setlength\tabcolsep{4.5pt}
\centering
\bgroup
\def\arraystretch{1.2}
\begin{tabular}{lccccccccccccc}
\toprule[0.8pt]
\multirow{2}{*}{Method} & \multicolumn{3}{c}{DuConv} & & \multicolumn{3}{c}{Persona-Chat} & & \multicolumn{3}{c}{CMU-DoG} \\
\cline{2-4} \cline{6-8} \cline{10-12}
 & F1$_{all}$ & F1$_{cross}$ & F1$_{intra}$ & & F1$_{all}$ & F1$_{cross}$ & F1$_{intra}$ & & F1$_{all}$ & F1$_{cross}$ & F1$_{intra}$ \\
\hline
All objectives & \textbf{88.60} & \textbf{83.72} & \textbf{89.27} & & {66.75} & \textbf{24.13} & {69.44} & & 58.45 & \textbf{58.92} & 58.82\\
\hdashline
\multicolumn{1}{l}{w/o TLM \& HPSI} & 88.07 & 81.90 & 89.06 & & 65.07 & 23.91 & 68.34 & & 58.23 & 53.15 & 59.24 \\
\multicolumn{1}{l}{w/o SPI \& UOR} & 87.75 & 81.56 & 88.81 & & \textbf{68.35} & 22.86 & \textbf{71.29} & & 58.08 & 47.93 & 60.22 \\
\multicolumn{1}{l}{w/o SAI} & 88.00 & 83.16 & 89.06 & & 64.74 & 23.33 & 67.99 & & \textbf{59.94} & 54.68 & \textbf{61.87} \\
only w/ TLM \& HPSI & 87.82 & 83.21 & 88.95 & & 65.56 & 24.12 & {68.60} & & 57.32 & 52.74 & 59.11 \\
only w/ SPI \& UOR & 88.45 & {83.70} & 89.10 & & 64.09 & 24.09 & 67.50 & & 59.71 & 57.23 & 60.80 \\
only w/ SAI & 88.49 & 82.97 & 89.24 & & 65.82 & 23.30 & 69.18 & & 57.20 & 50.54 & 57.63 \\
\hdashline
\multicolumn{1}{l}{w/ end2end pre-training} & 87.28 & 81.02 & 88.73 & & 64.37 & 21.17 & 67.77 & & 57.86 & 50.40 & 58.20 \\
\hline
\hline
\multicolumn{1}{l}{Ours$_{\text{XLM-R}}$} & \textbf{88.30} & \textbf{83.38} & \textbf{89.17} & & \textbf{65.57} & \textbf{24.11} & \textbf{68.51} & & \textbf{59.60} & \textbf{56.16} & \textbf{60.78}\\
\hdashline
\multicolumn{1}{l}{w/o SC-Encoder} & 88.02 & 79.11 & 89.05 & & 63.12 & 17.55 & 66.70 & & 57.72 & 50.42 & 58.03 \\
\multicolumn{1}{l}{w/o PA-Encoder} & 88.10 & 81.32 & 88.78 & & 64.05 & 22.38 & 64.82 & & 58.24 & 54.00 & 59.23 \\
\multicolumn{1}{l}{w/o SC-Encoder and PA-Encoder} & 86.14 & 73.63 & 87.12 & & 62.87 & 12.38 & 63.02 & & 52.44 & 41.02 & 56.23 \\
\hdashline
\multicolumn{1}{l}{w/o \textsc{MTrans}} & 88.25 & 83.01 & 89.08 & & 65.27 & 23.10 & 68.38 & & 58.58 & 55.41 & 59.98 \\
\toprule[0.8pt]
\end{tabular}
\egroup
\caption{Ablation studies on pre-training objectives and different modules.}
\label{tab:ablation-results}
\vspace{-0.5cm}
\end{table*}

\subsection{Main Results}
Table \ref{tab:csrl-results} summarized the results of all compared methods on DuConv, Persona-Chat and CMU-DoG datasets. Firstly, we can see that our method achieves competitive performance over all datasets, especially in cross-lingual scenario where our method outperforms the baselines by large margins no matter fine-tuning or freezing the language model during the CSRL training stage. Although CSAGN exceeds our method on DuConv test set, it fails to work well in cross-lingual scenario. We think this is because it heavily relies on the rich features from the Chinese pre-trained language model and it is overfitting on the predicate-aware information. Superior to CSAGN, our model with the multilingual backbone achieves outstanding performance on both language in-domain and cross-lingual datasets. This observation is expected because (1) our model is language-agnostic which makes the cross-lingual transfer easier; (2) our model captures high-level conversational features in SC-Encoder, thus enhancing the capacities of the model to recognize cross-arguments; (3) rich semantic features are modeled by PA-Encoder, which would improve the capacities of the model to recognize intra-arguments.

Secondly, although our model has achieved good performance over all datasets, further improvements can be observed after incorporating the proposed pre-training objectives, especially when freezing the parameters of the language model.
Exceptionally, we find that the performance on the CMU-DoG dataset heavily drops after introducing the pre-training objectives, especially in terms of F1$_{intra}$. We think this is because the semantic argument spans in CoNLL-2012 are relatively different from those in CMU-DoG, thus leading to the vague boundary detection and performance drop.
To verify this assumption, we conduct an ablation study by removing SAI from the pre-training stage. Interestingly, we observe substantial improvements over F1$_{all}$ and F1$_{intra}$, suggesting that pre-training on CoNLL-2012 does hurt the performance on CMU-DoG.
Furthermore, we also find that fine-tuning all parameters leads to slightly better performance than freezing the language model during the CSRL training stage. This finding is consistent with the previous work \citep{conia2021unifying}.

Table \ref{tab:ablation-results} presents the results of ablation studies on pre-training objectives and different modules. For the pre-training objectives, we found that (1) removing TLM \& HPSI objective hurts the performance consistently but slightly; (2) SPI \& UOR objectives help the model to better identify the cross-arguments; (3) SAI objective helps to find intra-arguments on DuConv and Persona-Chat, but might hurt the F1$_{intra}$ score on CMU-DoG; (4) hierarchical pre-training is superior to end-to-end pre-training which simultaneously optimizes all auxiliary objectives.
We think this is because the end2end pre-training is extremely unstable and confuses the optimization process of the model.

For model components, we found that only removing one of the SC-Encoder, PA-Encoder or \textsc{MTrans} slightly affect the performance. However, the performance heavily decreases when SC-Encoder and PA-Encoder are both removed. We think the reason is that at least one module is needed to capture the high-level features on the top of the language model. We preserve these two modules in our model since they essentially learn different abilities, i.e., the ability of dialogue modeling and semantics modeling, which also makes our model more explainable.


\begin{table}[t]
\small
\centering
\begin{tabular}{ll}
\toprule[0.8pt]
{U1} & how many games did the colts win? \\
    {U2} & \underline{the Colts}$_\textbf{ARG0}$ finished with a 12-2 record. \\
    {Question} & who did they \underline{play}$_{\textbf{predicate}}$ in the playoffs? \\
    \textbf{\textcolor{red}{Question$^{\prime}$}} & {\textcolor{red}{who did the Colts play in the playoffs?}} \\
\toprule[0.8pt]
\end{tabular}
\caption{An example of question-in-context rewriting.}
\label{tab:rewrite-case}
\vspace{-2em}
\end{table}

\begin{table*}[t!]
\fontsize{9}{10} \selectfont
\setlength\tabcolsep{4.5pt}
\centering
\bgroup
\def\arraystretch{1.2}
\begin{tabular}{lccccccccccc}
\toprule[0.8pt]
\multirow{2}{*}{Method} & \multicolumn{3}{c}{Persona-Chat (en)} & & \multicolumn{2}{c}{BConTrast (de)} & & \multicolumn{2}{c}{BSD (ja)} \\
\cline{2-4} \cline{6-7} \cline{9-10}
 & B1/2 & D1/2 & Human [1-5] & & B1/2 & D1/2 & & B1/2 & D1/2 \\
\hline
Seq2Seq & 0.138/0.069 & 0.051/0.094 & 2.72 & & 0.089/0.042 & 0.041/0.089 & & 0.125/0.051 & 0.123/0.248\\
\hdashline
mUniLM$_{\text{wo/CSRL}}$ & 0.188/0.113 & 0.114/0.217 & 3.02 & & 0.107/0.061 & 0.079/0.187 & & 0.162/0.080 & 0.175/0.320\\
mUniLM$_{\text{w/CSRL}}$ & {0.195}/{0.122} & {0.116}/{0.223} & {3.16} & & 0.112/0.065 & 0.082/0.191 & & 0.178/0.088 & 0.177/0.326\\
\hdashline
mBART$_{\text{wo/CSRL}}$ & 0.198/0.125 & 0.120/0.228 & 3.20 & & 0.115/0.072 & 0.086/0.206 & & 0.193/0.097 & 0.182/0.340\\
mBART$_{\text{w/CSRL}}$ & \textbf{0.217}/\textbf{0.136} & \textbf{0.124}/\textbf{0.233} & \textbf{3.25} & & \textbf{0.118}/\textbf{0.077} & \textbf{0.090}/\textbf{0.212} & & \textbf{0.205}/\textbf{0.110} & \textbf{0.185}/\textbf{0.346}\\
\toprule[0.8pt]
\end{tabular}
\egroup
\caption{Evaluations on response generation tasks in English, German and Japanese.}
\label{tab:response-results}
\vspace{-0.5cm}
\end{table*}

\subsection{Applications}
\citet{xu2021conversational} has confirmed the usefulness of CSRL by applying CSRL parsing results to two Chinese dialogue tasks, including dialogue context rewriting and dialogue response generation. In the same vein, we also explore whether CSRL could benefit to the same non-Chinese dialogue tasks.

\paragraph{Question-in-context Rewriting} \label{sec:rewrite}

\emph{Question-in-context rewriting} \citep{elgohary-etal-2019-unpack} is a challenging task which requires the model to resolve the conversational dependencies between the question and the context, and then rewrite the original question into independent one. This is an example in Table \ref{tab:rewrite-case}. The question {``who did they play in the playoffs?"} cannot be independently understood without knowing ``they'' refer to, but it can be resolved with the given context.

Since the CSRL models can identify the predicate-argument structures from the entire conversation, we believe that it can help this rewriting task by searching the dropped or referred components from the context. For example, in Table \ref{tab:rewrite-case}, our CSRL parser can find that the {ARG0} of the predicate ``play" is ``the Colts". Motivated by this observation, we attempt to borrow CSRL to help the question rewriting with the context.
We first employ the pre-trained cross-lingual CSRL parser (Ours$_{\text{XLM-R + pre-train}}$) to extract predicate-argument pairs from conversations.
We adopt the model proposed in \citep{xu2020semantic} to achieve the rewriting.
More details about the model are in Appendix \ref{apx:models}.
\begin{table}[t!]
\small
\centering
\begin{tabular}{lccc}
\toprule[0.8pt]
Method & B1 & B2 & B4 \\
\hline
Seq2Seq & - & - & 49.67\\
SARG\citep{huang2020sarg} & - & - & \textbf{54.80}\\
RUN\citep{liu2020incomplete} & \textbf{70.50} & 61.20 & 49.10 \\
Human evaluation & - & - & 59.92\\
\hline
Ours$_{\text{wo/~CSRL}}$ & 69.24 & 62.93 & 52.78 \\
Ours$_{\text{w/~CSRL}}$ & {70.26} & \textbf{64.19} & {54.23}\\
\toprule[0.8pt]
\end{tabular}
\caption{Evaluation results on the dataset of CANARD.}
\label{tab:canard-results}
\vspace{-2em}
\end{table}

Since the rewriting datasets are only available in Chinese and English, we hereby only evaluate on CANARD \cite{elgohary-etal-2019-unpack} which is a widely used English question rewriting dataset, and report the BLEU scores. Table \ref{tab:canard-results} lists the evaluation results on CANARD.
We can see that our implementation with CSRL achieves competitive performance against the state-of-the-art rewriting models, i.e., SARG \cite{huang2020sarg} and RUN \cite{liu2020incomplete}, and significantly outperforms the baseline method \citep{bahdanau2014neural}. Note that, in this part, we are more focused on the improvements after introducing CSRL information. We find that the scores across all metrics are improved with the aid of CSRL.
To figure out the reasons of these improvements, we investigate which type of questions could benefit from CSRL information most. By comparing the rewritten questions of different methods, we find that the questions that require information completion, especially those containing referred components (around 15\% cases), benefit from CSRL most. This observation is in line with our expectation that our CSRL parser could consistently offer essential guidance by recovering the dropped or referred text components.

\paragraph{Multi-turn Dialogue Response Generation}
Besides the rewriting task that is heavily affected by omitted components, we also explore the usefulness of CSRL to \emph{multi-turn dialogue response generation}, one of the main challenges in dialogue community.
In contrast to single-turn dialogue response generation, multi-turn dialogues suffer more frequently occurred ellipsis and anaphora, which leads to vague context representations.
To this end, we attempt to employ CSRL to build better context representations. In specific, we highlight the words picked up by the CSRL parser, and then teach the model to pay more attention on those words which would hold more semantic features.

We evaluate on three dialogue datasets in different languages, including \textbf{Persona-Chat} \citep{zhang2018personalizing} in English, \textbf{BConTrast} \citep{farajian2020findings} in German and \textbf{BSD} \citep{rikters2019designing} in Japanese. We report BLEU-1/2 and Distinct-1/2 scores for the comparison. We employ the pre-trained cross-lingual CSRL parser (Ours$_{\text{XLM-R}}$) to analyze the latest utterance, and obtain the predicate-argument pairs. Then the concatenated sequence of the extracted pairs and the context is fed into our model for response generation. We adopt the UniLM \citep{dong2019unified} or mBART \citep{liu2020multilingual} as our generation model. More implementation details are in Appendix \ref{apx:models}.

Table \ref{tab:response-results} summarizes the results on three datasets. We can see that the models with different backbones can consistently benefit from the additional introduced CSRL information. While substantial gains from CSRL information are obtained on English and Japanese dialogues, smaller improvements are observed on the German dialogue task. We think this is because English is well-represented in pre-trained multilingual models and Japanese is more similar to Chinese while German accounts for none of both. Apart from automatic evaluation criteria, we also conduct human evaluation on the English dataset. Specifically, we randomly select 200 generated responses for each method, and then recruit three annotators to evaluate the coherence and informativeness of the response against the conversation context by giving a score ranging from 1(worst) to 5(best).
We find that the method with CSRL wins in 35\% cases, and ties with the vanilla model in around 55\% cases.
With more careful analysis, we find that the responses that contains entities mentioned in histories benefit from CSRL information most. We think this is because none-phrases are more likely to be recognized as semantic arguments by CSRL parser, and then receive more attentions during encoding.

\section{Conclusion}
In this work, we propose a simple but effective model with five pre-training objectives to perform zero-shot cross-lingual CSRL,
and also confirm the usefulness of CSRL to non-Chinese dialogue tasks.



\bibliography{anthology,custom}
\bibliographystyle{acl_natbib}

\clearpage
\appendix
\section{Hard Parallel Sentence Identification Sampling} \label{apx:hpsi}
Following previous work \citep{robinson2020contrastive, wei2020learning} which suggests that contrastive learning of representations benefits from hard negative samples, we also try to select hard negative samples for PSI task based on n-gram similarity and text perturbation.
Specifically, for each sentence, we calculate its n-gram similarity scores to other sentences, where $n=1,2,3,4$, and then we select the sentence with the highest score at each gram as the candidate sentence; additionally, we construct the corrupted sentence as the candidate by token deletion, token replacement and token order permutation. Finally, we sample from the candidate set created by n-gram similarity at 40\% time and from the candidate set created by text perturbation at 60\% time.

\section{Modified Transformer Encoder Layer} \label{apx:mtrans}
To overcome the information forgetting of hierarchical models, we attempt to modify the standard Transformer to better reserve the information from the previous layers. In specific, we try following variants:

\begin{itemize}
    \item \textbf{\textsc{MTrans}}. Replacing the \texttt{[Add]} operation in the \textbf{first} residual connection layer with \texttt{[Concat]}.
    \item \textbf{\textsc{later-MTrans}}. Replacing the \texttt{[Add]} operation in the \textbf{second} residual connection layer with \texttt{[Concat]}.
    \item \textbf{\textsc{both-MTrans}}. Replacing the \texttt{[Add]} operations in \textbf{both} the first and second residual connection layers with \texttt{[Concat]}.
\end{itemize}

Our intuition of substituting the summation with concatenation is that the residual layer with concatenation would introduce additional parameters, and we expect these additional parameters to retain more history information. As shown in Table \ref{tab:csrl-results}, we obtain some gains while using \textsc{MTrans}. Additionally, we also report the F1$_{all}$ scores on DuConv/Persona-Chat/CMU-DoG datasets while using \textsc{later-MTrans} and \textsc{both-MTrans} here. \textsc{later-MTrans} achieves 88.18/65.32/58.44 points, and \textsc{both-MTrans} achieves 88.40/66.12/59.72 points against the standard Transformer achieving 88.25/65.27/58.58 points. Although \textsc{both-MTrans} achieves the best performance, we finally choose \textsc{MTrans} since \textsc{both-MTrans} brings a large volume of additional parameters which leads to a huge model size while the increasing of model parameters caused by \textsc{MTrans} is acceptable.

\section{Experimental settings}
We implement the model in PyTorch\citep{paszke2019pytorch}, and use the pre-trained language model of multilingual BERT (mBERT) or XLM-RoBERTa (XLM-R) made available by the Transformer library \citep{wolf2020transformers} as the backbone.
We train the model using AdamW\citep{loshchilov2018decoupled} with a linear learning rate schedule. For each model, we run five different random seeds and report the average score. More details and hyper-parameters are listed in Table \ref{tab:csrl-para}.

\section{Baselines} \label{apx:baselines}

We compare to following baseline models,

\begin{enumerate}
    \item \textbf{SimpleBERT/SimpleXLMR} \citep{shi2019simple}. It uses the Chinese BERT or XLM-R as the backbone and simply concatenates the entire dialogue context with the predicate.
    \item \textbf{CSRL- BERT/XLMR} \citep{xu2021conversational}. It uses the Chinese BERT or XLM-R as the backbone but attempts to encode the conversation structural information by integrating the dialogue turn and speaker embeddings in the input embedding layer.
    \item \textbf{CSAGN/CSAGN-XLMR} \citep{wu-etal-2021-csagn}. It uses the Chinese BERT or XLM-R as the backbone and employ the relational graph neural network to model predicate- and speaker-aware dependencies. We implement this baseline based on the code \url{https://github.com/hahahawu/CSAGN}.
\end{enumerate}

\section{Application Models} \label{apx:models}
\paragraph{Rewriting Model.} We adopt the model proposed in \citep{xu2020semantic} which directly concatenates the predicate-argument structures, the conversation context and the question as a sequence, and then feeds them into the model with special attention masks. During decoding, the model takes CSRL pairs and the context to generate the rewritten question word by word. The input representation, attention strategies and loss function of our model are same as \cite{xu2020semantic}'s. We initialize the model using the base BERT model and use AdamW with a linear learning rate schedule to update parameters.

Note that we only attempt to introduce the CSRL information as a condition into our generation-based model. We did not include the CSRL information into the state-of-the-art rewriting models, i.e., SARG and RUN because these models rewrite the sentence by learning a text editing matrix instead of directly learning the distributions of the target words. Unfortunately, there are no straightforward ways to include our CSRL information into these models to help the matrix learning.

\paragraph{Response Generation Model.} Our model for response generation is directly borrowed from UniLM \citep{dong2019unified} or mBART \citep{liu2020multilingual}.
For UniLM, the generation process is same with the rewriting task, wherein the extracted semantic pairs, the context and the response are concatenated into a sequence and encoded with the special mask. For mBART, we just concatenate the extracted predicate-argument pairs with the context into a sequence, and then feed the sequence into the encoder for training; during decoding, our model takes semantic information and the context as input to generate the response word by word. The input representation, attention strategies for CSRL structures and loss function are same as the rewriter model's. We initialize the model using the base multilingual BERT or mBART and use AdamW with a linear learning rate schedule to update parameters.


\section{Hyper-parameters} \label{apx:para}
We list the hyper-parameters of CSRL experiments (Table \ref{tab:csrl-para}), rewriting experiments (Table \ref{tab:rewrite-para}) and response experiments (Table \ref{tab:response-para}) below.

\begin{table}[ht!]
\small
    \fontsize{10}{11} \selectfont
    \setlength\tabcolsep{2.1pt}
    \centering
    \bgroup
    \def\arraystretch{1.2}
    \begin{tabular}{lr}
    \toprule
        Name & Value \\
        Language model & xlm-roberta-base \\
        Hidden state size & 512 \\
        Word-level encoder layers & 2 \\
        Pred.-arg encoder layers & 1\\
        Batch size per GPU & 24 \\
        Max learning rate & 5e-5 \\
        Min learning rate & 1e-5 \\
        Max \textit{lr} for LM fine-tuning & 1e-5 \\
        Min \textit{lr} for Lm fine-tuning & 1e-6 \\
        Max sequence length & 512 \\
        Max training epochs & 50 \\
        Max training steps & 15000 \\
        Early-stop patience & 10 \\
    \bottomrule
    \end{tabular}
    \egroup
    \caption{Hyper-parameters in CSRL experiments.}
    \label{tab:csrl-para}
\end{table}

\begin{table}[ht!]
\small
    \fontsize{10}{11} \selectfont
    \setlength\tabcolsep{2.1pt}
    \centering
    \bgroup
    \def\arraystretch{1.2}
    \begin{tabular}{lr}
    \toprule
        Name & Value \\
        Language model & bert-base-cased \\
        Hidden state size & 768 \\
        Batch size per GPU & 16 \\
        Max learning rate & 3e-5 \\
        Min learning rate & 1e-5 \\
        Max sequence length & 512 \\
        Max decode length & 32 \\
        Max training epochs & 20 \\
        Early-stop patience & 5 \\
    \bottomrule
    \end{tabular}
    \egroup
    \caption{Hyper-parameters in rewriting experiments.}
    \label{tab:rewrite-para}
\end{table}

\begin{table}[ht!]
\small
    \fontsize{10}{11} \selectfont
    \setlength\tabcolsep{2.1pt}
    \centering
    \bgroup
    \def\arraystretch{1.2}
    \begin{tabular}{lr}
    \toprule
        Name & Value \\
        Language model & mBERT and mBART \\
        Hidden state size & 768 \\
        Batch size per GPU & 16 \\
        Max learning rate & 5e-5 \\
        Min learning rate & 3e-5 \\
        Max sequence length & 512 \\
        Max decode length & 64 \\
        Max training epochs & 20 \\
        Early-stop patience & 5 \\
    \bottomrule
    \end{tabular}
    \egroup
    \caption{Hyper-parameters in response generation experiments.}
    \label{tab:response-para}
\end{table}

\end{CJK*}
\end{document}